\documentclass[10pt,twocolumn,letterpaper]{article}

\usepackage{iccv}
\usepackage{times}
\usepackage{epsfig}
\usepackage{graphicx}
\usepackage{amsmath}
\usepackage{amssymb}
\usepackage{multirow}
\usepackage{hhline}
\usepackage{array}
\usepackage{booktabs}
\usepackage{algorithm}
\usepackage{algorithmic}
\usepackage{caption} 
\usepackage[breaklinks=true,bookmarks=false]{hyperref}

\iccvfinalcopy 

\def\iccvPaperID{****} 

\ificcvfinal\fi

\begin{document}

\title{\ A Strong Baseline for Vehicle Re-Identification}

\author{Su V. Huynh, Nam H. Nguyen, Ngoc T. Nguyen, Vinh TQ. Nguyen, Chau Huynh, Chuong Nguyen\\
Cybercore AI\\
{\tt\small \{su.huynh, nam.nguyen, ngoc.nguyen, vinh.nguyen, chau.huynh, chuong.nguyen\}@cybercore.co.jp}
}

\maketitle
\ificcvfinal\thispagestyle{empty}\fi

\begin{abstract}
Vehicle Re-Identification (Re-ID) aims to identify the same vehicle across different cameras, hence plays an important role in modern traffic management systems. The technical challenges require the algorithms must be robust in different views, resolution, occlusion and illumination conditions. In this paper, we first analyze the main factors hindering the Vehicle Re-ID performance. We then present our solutions, specifically targeting the dataset Track 2 of the 5th AI City Challenge, including (1) reducing the domain gap between real and synthetic data, (2) network modification by stacking multi heads with attention mechanism, (3) adaptive loss weight adjustment. Our method achieves \textbf{61.34}\% mAP on the private CityFlow testset without using external dataset or pseudo labeling, and outperforms all previous works at \textbf{87.1}\% mAP on the \textit{Veri} benchmark. The code is available at 
 \url{https://github.com/cybercore-co-ltd/track2_aicity_2021}.

\end{abstract}
\section{Introduction}

Vehicle Re-ID aims to re-target vehicle images across non-overlapping camera views given a query image. It has many practical applications, such as for analyzing and managing the traffic flows in Intelligent Transport System.

Despite many progresses have been made in the recent years thanks to deep learning, vehicle Re-ID is still facing many challenges, such as severe variations from different view points, partial occlusion, image blurry or illumination changes. The state-of-the-art methods \cite{1st, 2nd, 3rd} typically use a deep neural network to extract the vehicle visual representation. Some methods proposed to enhance the feature representation by using multi-head architecture to extract multi-scale information, such as 
Zheng {\it et al.} \cite{1st}. However, they only use simple pooling operators to extract feature vectors, which then be averaged in inference stage. Hence, the feature lacks the vehicle detailed characteristics, which is important to distinguish objects with similar appearance. 
\begin{figure}[H]
	\includegraphics[width=\linewidth]{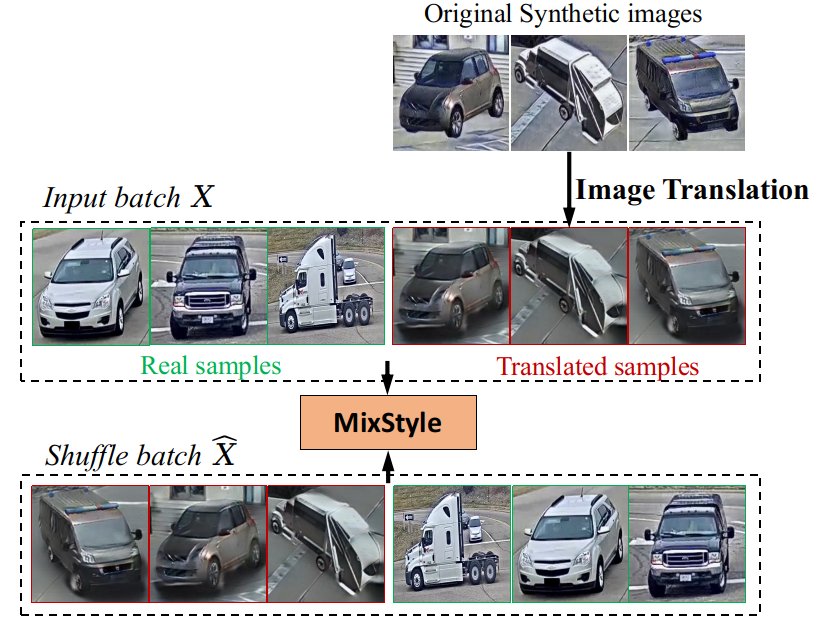}
	\caption{Domain Generalization with MixStyle.}
	\label{fig:mixstyle}
\end{figure}
\begin{figure*}
	\centering
	\includegraphics[width=\linewidth]{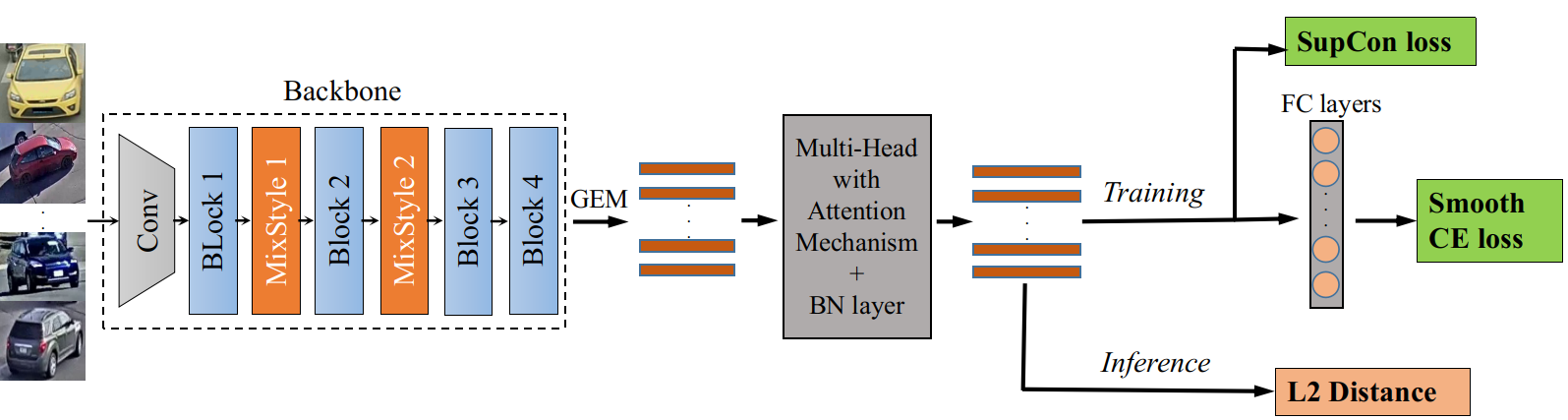}
	\caption{The training pipeline. GEM: Generalized Mean Pooling, BN: Batch Normalization, FC: Fully Connected, CE: Cross Entropy, SupCon: Supervised Contrastive.}
	\label{fig:training}
\end{figure*}
In contrast, in Face ID application, Kim {\it et al.} \cite{groupface} propose to retain the detailed characteristics into several latent groups, which improve the Re-ID confidence. 
This motivates us to develop a mechanism to integrate the feature vectors into several common groups, which help filtering out the candidates during retrieval. 
To improve the model generalization, a typical practice is to train the model on large datasets, such as the Veri dataset \cite{veri} and PKU-VehicleID dataset \cite{pku-dataset}. Another approach to obtain large dataset without costly human labeling is to utilize synthetic data generated from 3D simulation environment, by which we can have fully control of the vehicle's appearance \cite{vehicleX}. In the Track 2 of the 5th AI City Challenge, two datasets are provided, namely the real-world and synthetic data, as illustrated in Figure \ref{fig:mixstyle}. However, as seen in Figure \ref{fig:mixstyle}, there is always a domain gap between two data sources, which leads to feature distribution shifting. To tackle this problem, Zheng {\it et al.} \cite{1st} adopt the image translation technique (UNIT \cite{unit}) to transform the synthetic data closer to the realistic one. However, the translated images are still in poor quality, and the domain gap is still significant. 

In addition, designing effective loss functions to train the network is also very important. A majority of previous works \cite{zheng2020beyond, 1st, 2nd, 3rd} typically use a combination of Triplet Loss and Cross Entropy Loss. The loss function 's objective is intuitive: pulling samples with the same ID together, while pushing those with different IDs far apart. However, the Triplet Loss function only uses one
positive and one negative pair per anchor, and the hard negative mining process must be tuned carefully. Moreover, the ratio between Triplet Loss and Cross Entropy Loss is heuristically set to 1:1. However, our experiments show that this ratio setting also has a strong impact to the performance, but surprisingly, to the best of our knowledge, it is often overlooked in the previous works. 

From the aforementioned analysis, we present our solutions to adress the problem. Our main contributions are:

(1) We adopt MixStyle Transfer \cite{mixstyle} as a regularization method to reduce the gap between the real and synthetic data.

(2) Multi-head with attention mechanism are attached to the backbone to help the model learn more detailed features. The features are then automatically grouped into sub-features, each help narrows down the search space of the target identity.

(3)  We replace the commonly used Triplet Loss with the Supervised Contrastive Loss \cite{supcon} which help the network learning more effectively. Additionally, a novel adaptive loss weight between the Supervised Contrastive Loss and the Cross Entropy Loss is provided to improve the performance dramatically.

\section{Related work}

In order to enrich visual representation for deep learning based models, a large scale dataset is necessary. Liu {\it et al.} \cite{veri} propose the VeRi dataset, which contains a large number of vehicles captured by non overlapping cameras with different perspectives, scales and illuminations in real world urban traffic. In addition, annotating dataset is very costly and time consuming. To solve this problem, many efforts have been made to improve the data generation techniques. For instance, Yao {\it et al. } \cite{vehicleX} introduce a large-scale synthetic dataset simulated by a flexible 3D graphic engine with editable attributes such as vehicle orientation, light direction and camera height. Recently, the generative adversarial network (GAN) can be used to generate new data by transferring vehicle style \cite{Deng_2018_CVPR, 8578114} or changing the vehicle attributes \cite{zheng2019joint}. Moreover, Zhou {\it et al.} propose the MixStyle method \cite{mixstyle}, which attempts to create a domain-generalized model by mixing the feature statistics to simulate new styles. However, simply adding synthetic data to train the model often yeilds inferior results, due to the domain gap and feature bias between the synthetic and the real world data.

Other methods focus on developing more effective loss functions to improve the network training efficiency. For instance, the Large Margin Cosine Loss \cite{cosface} aims to maximize the inter-class variance and minimize intra-class variance. The Triplet Loss \cite{triplet} aims to learn visual representation by optimizing the distances between a set of three hard samples. Sun {\it et al.} \cite{circleloss} propose the Circle Loss, which adaptively adjusts weights for each similarity score. In addition to loss functions, sampling strategy also plays an important role in re-id training. The Hierarchical Triplet Loss \cite{HTL} uses a predefined hierarchical tree to formulate informative training samples, which help to overcome the limitation of random sampling when training triplet loss. The semi-hard triplet mining \cite{facenet} focuses on negative examples which have close distances to the anchor positive distances. However, sampling strategy is generally heuristic, depending on the loss function, and hard to tune.

Additionally, post-processing is also important to reduce the false-positive prediction. For example, re-ranking can improve the accuracy of the ranking list. Re-ranking approaches are widely used in person re-id \cite{old_reranking_1}, \cite{old_reranking_2}, which typically rely on the consistency and nearest-neighbor relationship of gallery images based on initial re-id ranking. Recently, Zhong {\it et al.} \cite{rerank} propose the k-reciprocal encoding method , which considers the original distance and the Jaccard distance between two images. In this work, we also perform an ablation study to find the best practice in applying post-processing steps to the vehicle Re-ID problem. 
\section{Proposed Method}

In section \ref{sub:domain_generalization}, we introduce the algorithm to bridge the gap between synthetic data and real data. Then we show our baseline architecture which applies multi-head with attention mechanism in section \ref{sub:network_architecture}. In section \ref{sub:loss_function}, the alternative Contrastive Loss and Adaptive Loss Weight are introduced. Finally, we present some bag of post-processing tricks in section \ref{sub:post_processing}.

\subsection{Domain Generalization} \label{sub:domain_generalization}
To train a model that generalizes to unseen domains, we adopt the MixStyle method \cite{mixstyle}, which aims to simulate new styles by mixing the statistical features of two samples from different domains. Given the input batch $X$ (i.e.,  real and synthetic samples in a same batch training) and a shuffle of $X$, named $\hat{X}$, MixStyle computes the mixed feature's statistics by

\begin{equation}
\mu_m = \lambda\mu(X)+(1-\lambda)\mu(\hat{X})
\end{equation}
\begin{equation}
\sigma_m = \lambda\sigma(X)+(1-\lambda)\sigma(\hat{X})
\end{equation}
where $\lambda$ is the weights sampled from $Beta$ distribution, $\lambda$ $\sim$ $Beta$($\alpha$, $\alpha$). Following \cite{mixstyle}, we set $\alpha=0.1$ throughout all our experiments. Rely on the mixed feature statistic, style-normalized $X$ is computed as 
\begin{equation}
\rm{MixStyle}(X)=\sigma_m\frac{X-\mu(X)}{\sigma(X)} + \mu_m .
\end{equation}

By leveraging the feature-level statistics, MixStyle implicitly regularizes the network. This makes the model become more robust to the domain difference and enforce the network to learn the object semantic features.

\subsection{Network Architecture} \label{sub:network_architecture}
We adopt the method proposed in \cite{2nd} as the baseline, and augment it with our proposed network modification. The network architecture, the training and inference pipeline are illustrated in Figure.\ref{fig:training}.

\paragraph{Backbone.} We use Instance Batch Normalization (IBN) network family \cite{ibn} as the backbone due to its advantages. Firstly, by utilizing the instance normalization, the feature extractor can learn robust encoded representations that invariant to appearance differences. Secondly, it can improve the performance of other advanced deep neural network architecture such as ResNet, ResNeXt, and SENet. Moreover, we attempt to append MixStyle layers into the network to improve the domain generalization. Specifically, as described in \cite{mixstyle}, convolution layers in early stages encode the style information, while later stages tend to capture the semantic content. Therefore, we add the MixStyle module after the Block 1 and 2 in the ResNeXt\_ibn\_a 101 backbone \cite{ibn}, as shown in Figure \ref{fig:training}. 

\begin{figure}
	\includegraphics[width=\linewidth]{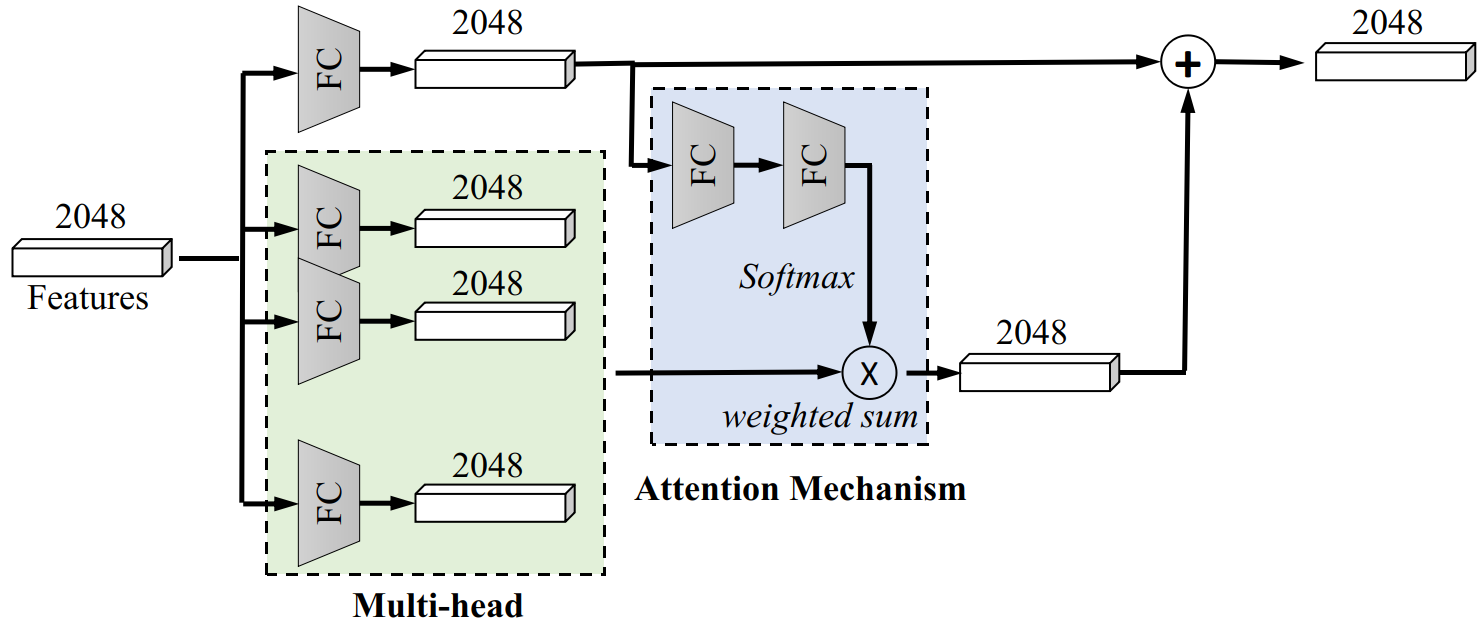}
	\caption{Multi-head with attention mechanism.}
	\label{fig:attention}
\end{figure}

\paragraph{Multi-head with Attention Mechanism.} Distinguishing thousands of vehicles with multiple views is challenging. Instead of using only one head, using multi-head encourages the re-id model to learn more diverse features from different vehicle characteristics. Thus, we adopted the multiples heads architecture \cite{groupface} to further enhance the quality of the visual representation for vehicle re-id. Figure \ref{fig:attention} shows the architecture of the multiple head with attention mechanism. In particular, the 2048-dim feature obtained from the backbone is fed into multiple parallel fully connected (FC) layers. Following \cite{groupface}, each FC layer is considered as one head and expected to learn distinct features which take into account different vehicle characteristics. Additionally, the attention mechanism determines which head's features are more important to the final encoding feature. 

\subsection{Loss Function} \label{sub:loss_function}
\subsubsection{ID and Metric Losses } \label{sub:id_metric_loss}
For training re-id model, a common approach is to use a combination of \textit{ID Loss} and \textit{Metric Loss}. In particular, Cross Entropy (CE) is used for ID Loss to classify samples in different classes, while Metric Loss is often the Contrastive Loss, such as Triplet Loss \cite{triplet} or Circle Loss \cite{circle_loss}, to optimize the feature distance between each class.

\paragraph{ID Loss.} In this work, we use the CE Loss for the ID Loss. Given an input image, the ID embedding vector is extracted from the fully connected layer attached to the multi-head module with the dimension equal to the number of vehicles $N$. Let $y$ is the ground truth ID label and $p_i$ is the ID prediction logits of class $i$, we use the Label Smoothing technique \cite{lsce} to prevent the model from over-fitting and improve robustness, the Label Smoothing CE Loss is defined as:

\begin{equation}
	\mathcal{L}(ID) = \sum_{i=1}^{N}-q_i log (p_i) 
	\begin{cases}
		q_i = 0, y \neq i \\
		q_i = 1, y = i
	\end{cases}
\end{equation}
\begin{equation}
	q_i =
	\begin{cases}
		1 - \frac{N-1}{N} \varepsilon       & \quad \text{if } i = y\\
		\varepsilon / N  & \quad \text{otherwise, } 
	\end{cases}
\end{equation}
where, $\varepsilon$ is a soft-margin to reduce the model over-confidence and is set to 0.1 in our experiments.	\\\\
\textbf{Metric Loss}. To improve the model performance on hard samples, we adopt the Supervised Contrastive Loss (SupCon) \cite{supcon}. Specifically, the SupCon can be seen as a generalized case of the Triplet and N-pair loss. Instead of using only one positive and one negative pair for each anchor, the SupCon considers many positive and negative pairs. Applying SupCon to the ReID problem provides several benefits (1) the gradient of SupCon loss function encourages learning from hard positives and hard negatives; and (2) it is less sensitive to hyper-parameters. The SupCon is computed as:

\begin{equation}
	\mathcal{L} = \sum_{i\in I} \frac{-1}{|P(i)|} \sum_{p \in P(i)} log \frac{exp(z_i z_p / \tau)}{\sum_{a \in A(i)} exp(z_i z_a / \tau)},
\end{equation}
where $P(i) \equiv \{p \in A (i) : \boldsymbol{\tilde{y}}_p = \boldsymbol{\tilde{y}}_i\}$ is the set of indices of all positives in the multi viewed batch distinct from i, $|P(i)|$ is its cardinality, $\tau \in \mathcal{R^+}$ is a scalar temperature parameter, $z_i$ is an anchor feature, $z_p$ is a positive feature and $z_a$ is a negative feature.

\subsubsection{Constructing Adaptive Loss Weight } \label{sub:calw}
\begin{figure}
	\includegraphics[width=\linewidth]{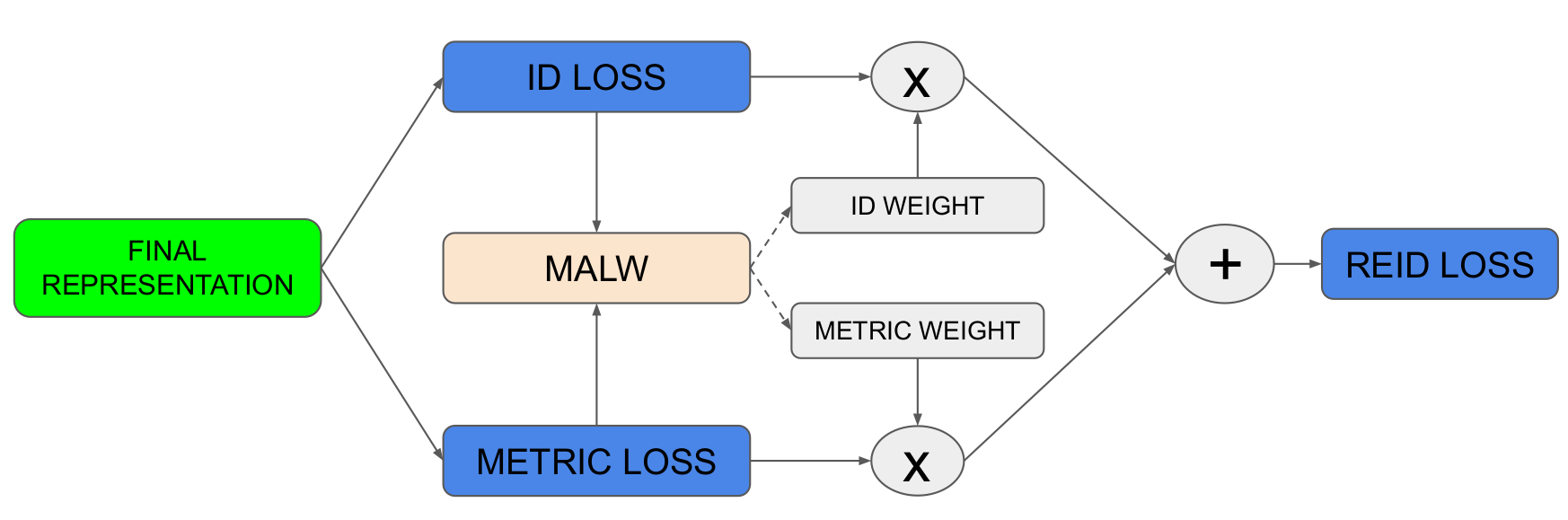}
	\caption{Momentum Adaptive Loss Weight (MALW).}
	\label{fig:malw}
\end{figure}

\textbf{Problems in Training}. Training the ReID model requires optimizing a combination of ID Loss and Metric Loss. Conventionally, the loss weights are set equally, i.e. 1:1 ratio. However, in practice, ID Loss is relatively much larger than Metric Loss, which causes the imbalance and affects the training performance. Table \ref{tab:lossweight} shows the sensitiveness of performance towards loss weight. Unfortunately, manually tuning the loss weight is sub-optimal and time consuming. Hence, motivated by the Adaptive Loss Weight Adjustment\cite{awla}, we propose the Momentum Adaptive Loss Weight (MALW) to increase training stability by automatically updating loss weights according to the statistical characteristics of loss values.	
\begin{table}[H]
	\centering
	\begin{tabular}{c| c c c c} 
		\hline
		Loss weight & 1:1 & 1:2 & 0.5:0.5 &MALW \\ [0.5ex] 
		\hline\hline
		mAP(\%) & 73.3 & 75.2 & 76.8 &\textbf{78.4} \\ 
		\hline
	\end{tabular}
	\caption{The performance of Baseline from \cite{2nd} under different loss weights (Cross Entropy Loss weight:Triplet Loss weight) and MALW.}
	\label{tab:lossweight}
\end{table}
\paragraph{Momentum Adaptive Loss Weight.}
Algorithm \ref{al:solution} and Figure \ref{fig:malw} describe how the MALW updates the weights during training progress. Let $\lambda_{ID}$ and $\lambda_{Metric}$ be the loss weights for ID Loss and Metric Loss, respectively. Initially, the ratio between $\lambda_{ID}$ and $\lambda_{Metric}$ is set to 1:1. After K iterations training, the ID loss weight $\lambda_{ID}$ is updated based on the standard deviation of the recorded ID Loss $L_{ID}$ and Metric Loss $L_{Metric}$ with a momentum factor. The MALW method improves our model performance by balancing the training losses without adding any computation cost to the inference step, as seen in Table \ref{tab:lossweight}.

\begin{algorithm}[t]
	\renewcommand{\algorithmicrequire}{\textbf{Input:}}
	\renewcommand{\algorithmicensure}{\textbf{Output:}}
	\caption{Momentum Adaptive Loss Weight}	\begin{algorithmic}[1]
		\REQUIRE
		{\begin{minipage}[t]{6cm}%
				\strut
				- ID Loss weight $\lambda_{ID}$ \\
				- Metric Loss weight $\lambda_{Metric}$ \\
				- Update iteration count $k$ \\
				- Momentum factor $\alpha$
				\strut
			\end{minipage}%
		}
		\STATE \textbf{Begin:}
		\STATE Initialize loss weight $\lambda_{ID}$, $\lambda_{Metric}$ to $1:1$
		\STATE Build two empty sets $S_{ID}$, $S_{Metric}$ for recording losses
		\FOR{$i=0$ to max iter}
		\STATE Obtain initial loss $L_{ID}$ and $L_{Metric}$
		\STATE Set $L_{ID} = \lambda_{ID} L_{ID}$, $L_{Metric} = \lambda_{Metric} · L_{Metric}$
		\STATE Add $L_{ID}$ to $S_{ID}$ and $L_{Metric}$ to $S_{Metric}$ 
		\IF{$i \% k == 0$}
		\STATE $ID_{std} = std(S_{ID})$, $Metric_{std} = std(S_{Metric})$
		\STATE Empty sets $S_{ID}$, $S_{Metric}$
		\IF{$ID_{std} > Metric_{std}$}
		\STATE $new\_\lambda_{ID} = 1 - (ID_{std} - Metric_{std}) / ID_{std}$
		\STATE Update weight: $\lambda_{ID} = \alpha * \lambda_{ID} + new\_\lambda_{ID}$
		\ENDIF
		\ENDIF
		\ENDFOR	
		\ENSURE $(\lambda_{ID}, \lambda_{Metric})$
	\end{algorithmic}
	\label{al:solution}
\end{algorithm}

\subsection{Post-Processing} \label{sub:post_processing}
\paragraph{Re-rank using k-reciprocal encoding.} We improve performance of the re-id model using a re-ranking method described in \cite{rerank}. This approach refines the initial ranking list using the information of original distance and Jaccard distance between two vehicle images.

\paragraph{Fused distance.} To reduce the influence of vehicle orientations and camera viewpoints, we adopt the fusion technique proposed in \cite{2nd}. Specifically, the vehicle ID, orientation and camera distance matrices are fused to get the cost fusion matrix, which is then used to find the optimal results for query images, as:
\begin{equation}
D(x_i, x_j) = D_v(x_i, x_j) - \lambda_1 D_o(x_i, x_j) - \lambda_2 D_c(x_i, x_j), 
\end{equation}
where $D_v(x_i, x_j)$, $D_o(x_i, x_j)$, $D_c(x_i, x_j)$ are ID distance, orientation distance and camera distance between two vehicle images $(x_i, x_j)$ , respectively.

\paragraph{Tracklet-Level Re-Ranking.} Additionally, we apply another re-ranking method using the tracklet information of vehicles which is included in the CityFlow dataset. To be specific, a vehicle 's tracklet is created from the detection and tracking results in one camera. Replacing features of each image in a tracklet with averaging the features of a subset of consecutive frames can help us enhance the visual representation of the same vehicle \cite{2nd}.

\paragraph{Ensemble.} We combined all 3 models using different backbones, including ResNet50\_ibn\_a \cite{ibn}, ResNeXt101\_ibn\_a \cite{ibn}, ResNet152 \cite{resnet}, by taking the averaged distance of each query image to gallery images. As shown in Table \ref{tab:posprocess}, our ensemble model significantly increases 2.8\% mAP on the CityFlow test set.
\section{Experimental Results}
\begin{table*}[h!]
	\centering
	\begin{tabular}{l|cc|cc}
		\hline
		Data & \multicolumn{2}{c}{Real-split} & \multicolumn{2}{c}{CityFlow}\\
		& mAP(\%)  & Rank 1(\%) & mAP(\%) & Rank 1(\%) \\
		\hline\hline
		Real & 75.5  & 79.7 & 22.1 & 31.6  \\
		\hline
		Real + Syn & 80.2  & 85.2 & 32.5 & 51.8  \\
		\hline
		Real + Syn (translated) & 81.5  & 86.7 & 35.3 & 54.3  \\
		\hline
		Real + Syn (translated) + Mixstyle & 83.8  & 88.7 & 37.7 & 58.2  \\
		\hline
	\end{tabular}
	\caption{Different datasets on Real-split and CityFlow.}
	\label{tab:dataset}
\end{table*}

\begin{table*}[h!]
	\centering
	\begin{tabular}{l|cc|cc}
		\hline
		Method & \multicolumn{2}{c}{Real-split} & \multicolumn{2}{c}{CityFlow}\\
		& mAP(\%)  & Rank 1(\%) & mAP(\%) & Rank 1(\%) \\
		\hline\hline
		Baseline + Multiple Head & 84.5 & 89.0  & 41.9 & 65.6  \\
		\hline
		Baseline + Multiple Head + SupCon & 85.7   & 90.9 & - & -  \\
		\hline
		Baseline + Multiple Head + SupCon + MALW & 88.1  & 92.5 & 49.5 & 58.2  \\
		\hline
	\end{tabular}
	\caption{Different training methods on Real-split and CityFlow.}
	\label{tab:method}
\end{table*}

\begin{table}
	\centering
	\begin{tabular}{l|cccc}
		\hline
		Method & \multicolumn{4}{c}{Performance} \\
		\hline\hline
		Re-rank & \checkmark  & \checkmark  & \checkmark  & \checkmark  \\
		Orientation \& Camera ID && \checkmark  & \checkmark  & \checkmark \\
		Track-rank ReID &&& \checkmark  & \checkmark \\
		Ensemble &&&& \checkmark \\
		\hline
		mAP(\%)  & 49.5  & 53.7  & 58.5 & 61.3 \\
		Rank 1(\%) & 58.2 & 64.7  & 70.2 & 72.2 \\
		\hline
	\end{tabular}
	\caption{Different pos-process techniques on CityFlow.}
	\label{tab:posprocess}
\end{table}

\subsection{Data Analysis}
The dataset of Track 2 challenge is the new version of CityFlow called CityFlowV2-ReID. There are 440 IDs retrieved from 52,717 images for training and other 440 identities come from 31.238 images in the test set. Following the restriction of using external data, our team tackles the problem of data limitation by leveraging the synthetic dataset called VehicleX \cite{vehicleX}. There are totally 1362 unique identities and 192150 synthetic images in VehicleX dataset, which can be used for model training or transfer learning. We split the real training data into Split-Train and Split-Test to validate offline. In particular, Split-Train contains 44375 images of 360 IDs, while Split-Test includes 8342 images of 80 IDs.
\subsection{Training Strategy}
We resize the images to (320x320) and apply several data augmentation methods, such as color jitters, random flip, brightness and contrast adjustment, random erase and random cropping. We use ADAM [11] optimizer with the cosine annealing scheduler and set the learning rate to 3.5e-4. The batch size is set to 128, which compose of 16 identities, where each identity contains 8 images. For vehicle re-id model, we adopt three strong backbones: ResNet50\_ibn\_a \cite{ibn}, ResNeXt101\_ibn\_a \cite{ibn} and ResNet152 \cite{resnet}  as our feature extractor. In addition, only ResNeXt101\_ibn\_a is used to train both Camera Re-ID and Orientation Re-ID models. All models are pretrained on ImageNet. For each single model, we first frozen the backbone and train multi heads for 1 epoch. Then we train the whole architecture for extra 11 epochs.
\subsection{Ablation Study}
In this section, we use Split-Train and Split-Test for ablation study. The results are summarized in Table \ref{tab:dataset}, \ref{tab:method}, and \ref{tab:posprocess}.

\paragraph{Baseline:} We use the standard backbone ResNeXt101\_ibn\_a \cite{ibn} along with CE Loss and Triplet Loss as a baseline for our vehicle re-id model, after training the baseline using the CityFlow dataset, we achieved 75.5\% mAP and 22.1\% mAP on Split-Test and CityFlow dataset.

\paragraph{Synthetic Data with MixStyle:} To evaluate the effectiveness of using synthetic data, we train the baseline using the combination of real and synthetic data, result in the increment of mAP to 80.2\% and 32.5\% on Split-Test and CityFlow, respectively. This indicates that proper usage of synthetic data to train the network is helpful. Moreover, by using translated synthetic data instead of the original one, the mAP increases to reach 81.5\% and 35.3\%. We further alleviate the domain gap between these two data sources by applying MixStyle, gaining 2.3\% and 2.4\% more mAP score on these datasets, as shown in Table \ref{tab:dataset}. This result opens up the possibility of using synthetic data for training deep re-id networks to reduce the cost of collecting real-word data and human annotation.

\paragraph{Multi-Head with Attention Mechanism:} After getting the data strategy for training, we enhance the network capability by applying Multi-head with Attention Mechanism and achieve 84.5\% and  41.9\% mAP on Split-Test and CityFlow, respectively, as shown in Table \ref{tab:method}. This demonstrates that the visual representation features obtained from multi-head are more robust compared to using only one single head.

\paragraph{Losses:} The combination of CE Loss and Triplet Loss is widely used in re-id tasks. Here, replacing the Triplet Loss by the Supervised Contrastive Loss \cite{supcon} results in the increment of 1.2\% mAP, from 84.5\% to 85.7\% on Split-Test set. Moreover, the MALW is applied to balance the loss functions, which solves the slow convergence problem and eliminate the need of loss weight setting. We set $K=500$ and $\alpha=0.9$ and this helps improve our mAP to 88.1\% and 49.5\% on Split-Test and CityFlow, respectively. The results are summarized in Table \ref{tab:method}.

\paragraph{Post-Processing:} Table \ref{tab:posprocess} illustrates the results of applying different post-processing methods. Firstly, the re-ranking algorithm \cite{rerank} is widely used and demonstrated its improvement, therefore, by default we apply it to all models. Secondly, the fused distance approach using the vehicle ID, Orientation and Camera distances \cite{2nd} increases mAP from from 49.5\% to 53.7\%, which indicates that the Orientation and Camera information is useful for the ReID performance. Thirdly, by applying track-ranking algorithm, we further gain 4.8\% mAP. Finally, after ensembling our three best single models, we achieve 61.34\% mAP on the CityFlow test set without using any external data or pseudo tricks.

\subsection{Performance on VeRi776}

To further demonstrate the generalization across datasets, we also test our proposed method on the Veri benchmark dataset. For a fair comparison, we only use single model, including backbone ResNeXt101\_ibn\_a, multi-head, CE and SupCon losses and MALW without applying pos-processing technique and synthetic data. We achieve the state-of-the-art performance with a large margin compared to previous works, as shown in Table \ref{tab:veri}.
\begin{table}
	\centering
	\begin{tabular}{l|cc}
		\hline
		Data & \multicolumn{2}{c}{Veri dataset} \\ & 
		mAP(\%) & Rank 1 (\%) \\
		\hline\hline
		Strong Baseline \cite{strongbaseline} & 67.6  & 90.2 \\
		DMML \cite{dmml}  & 70.1  & 90.2   \\
		PAMTRI(ALL) \cite{all} & 71.8  & 92.8  \\
		VOC ReID \cite{2nd} &82.8  & \textbf{97.6} \\
	
		\textbf{Our} &\textbf{ 87.1}  & 97.0 \\
		\hline
	\end{tabular}
	\caption{Comparison with the state-of-the art methods on the VeRi776 dataset.}
	\label{tab:veri}
\end{table}

\subsection{Visualization of results}
\begin{figure*}
	\centering
  \includegraphics[width=160mm]{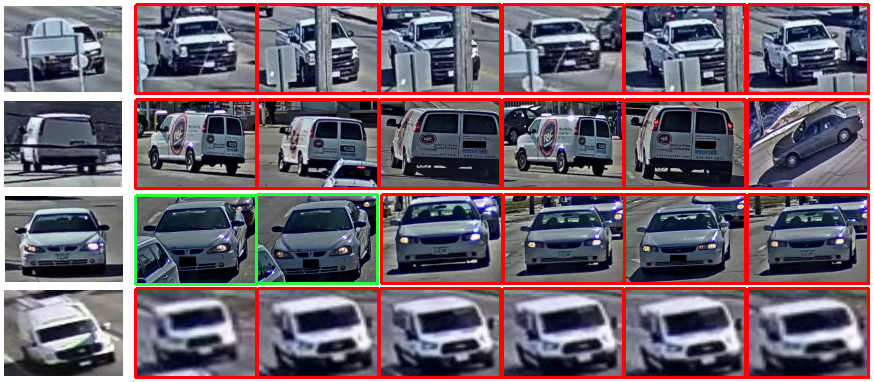}
  \caption{Result on the baseline model. Each row presents the query images and retrieved top 6 gallery images. Green and red boxes denote true positive and false positive sample, respectively.}
  \label{fig:baseline}
\end{figure*}
\begin{figure*}
	\centering
  \includegraphics[width=160mm]{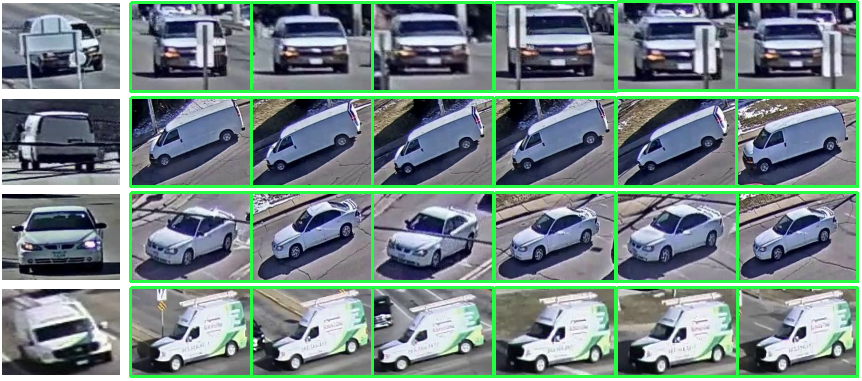}
  \caption{Result on the final model. Each row presents the query images and retrieved top 6 gallery images. Green and red boxes denote true positive and false positive sample, respectively.}
  \label{fig:final}
\end{figure*}

We visualize the query images and ranking lists obtained by the baseline model and our final model. As shown in Figure \ref{fig:baseline}, the baseline model fails to retrieve an accurate ranking list, since identifying vehicle objects from these query images is truly challenging. For example, the vehicle in the first row is occluded. Vehicles in the second and third row have similar appearance to other vehicles in the dataset, while the samples in the last row has very low resolution. On the contrary, our model surpasses the baseline and can retrieve a high quality ranking list, as shown Figure \ref{fig:final}.

\section{Conclusion}
In this paper, we proposed a strong baseline for the vehicle re-identification problem. By making improvements on utilizing the usage of real and synthetic data, employing the multi-head with attention mechanism and optimizing a combination of training losses, we achieve 61.34\% mAP on the CityFlow dataset. In the VeRi dataset, we achieve 87.1\% mAP, outperform the previous works with a large margin. Our method is simple, and focuses on improving the training techniques more efficiently.
Hence, it can be generally applied to a variety of Re-ID problems. We also released the code to facilitate the reproduction, hoping that it can serve a new baseline for further research. 
{\small
	\bibliographystyle{ieee_fullname}
	\bibliography{egbib}
}
\end{document}